\pdfoutput=1

\documentclass[11pt]{article}

\usepackage[final]{acl}

\usepackage{times}
\usepackage{latexsym}

\usepackage[T1]{fontenc}
\usepackage{CJKutf8}

\usepackage[utf8]{inputenc}

\usepackage{microtype}

\usepackage{inconsolata}

\usepackage{graphicx}

\usepackage{adjustbox}

\usepackage{tabularx}
\usepackage{booktabs}
\usepackage{multirow}
\usepackage{makecell}
\usepackage{fancyvrb,fvextra}

\usepackage{subfig}
\usepackage{tikz}
\usetikzlibrary{shapes.geometric, arrows, positioning}

\tikzstyle{data} = [rectangle, rounded corners, minimum width=2cm, minimum height=1cm, draw=black, fill=red!30]
\tikzstyle{qa} = [rectangle, rounded corners, minimum width=2cm, minimum height=1cm, draw=black, fill=gray!30]
\tikzstyle{articles} = [cylinder, shape border rotate=90, aspect=0.3, minimum width=1.5cm, minimum height=1cm, draw=black, fill=blue!20]
\tikzstyle{arrow} = [thick,->,>=stealth]
\tikzstyle{process} = [circle, minimum size=0.8cm, draw=black, fill=white]

\usepackage{xurl}

\usepackage{dsfont}
\usepackage{amsfonts}
\usepackage{amsmath}
\usepackage{stackrel}

\usepackage{listings}
\usepackage{here}

\usepackage[english]{babel}

\newcommand{\jp}[1]{\begin{CJK}{UTF8}{ipxm}#1\end{CJK}}

\title{Do Large Language Models Know Folktales? \\
A Case Study of Yokai in Japanese Folktales}

\author{Tsutsumi Ayuto \\
  Tokyo Metropolitan University \\
  \texttt{tsutsumi-ayuto@ed.tmu.ac.jp} \\\And
  Yuu Jinnai \\
  CyberAgent  \\
  \texttt{jinnai\_yu@cyberagent.co.jp} \\}

\begin{document}
\maketitle
\begin{abstract}
Although Large Language Models (LLMs) have demonstrated strong language understanding and generation abilities across various languages, their cultural knowledge is often limited to English-speaking communities, which can marginalize the cultures of non-English communities. 
To address the problem, evaluation of the cultural awareness of the LLMs and the methods to develop culturally aware LLMs have been investigated.
In this study, we focus on evaluating knowledge of folktales, a key medium for conveying and circulating culture.
In particular, we focus on Japanese folktales, specifically on knowledge of \textit{Yokai}.
Yokai are supernatural creatures originating from Japanese folktales that continue to be popular motifs in art and entertainment today. Yokai have long served as a medium for cultural expression, making them an ideal subject for assessing the cultural awareness of LLMs.
We introduce YokaiEval, a benchmark dataset consisting of 809 multiple-choice questions (each with four options) designed to probe knowledge about yokai. We evaluate the performance of 31 Japanese and multilingual LLMs on this dataset. The results show that models trained with Japanese language resources achieve higher accuracy than English-centric models, with those that underwent continued pretraining in Japanese, particularly those based on Llama-3, performing especially well. The code and dataset are available at \url{https://github.com/CyberAgentAILab/YokaiEval}.
\end{abstract}

\section{Introduction}

\begin{figure}[t]
    \centering
    \includegraphics[width=0.95\linewidth]{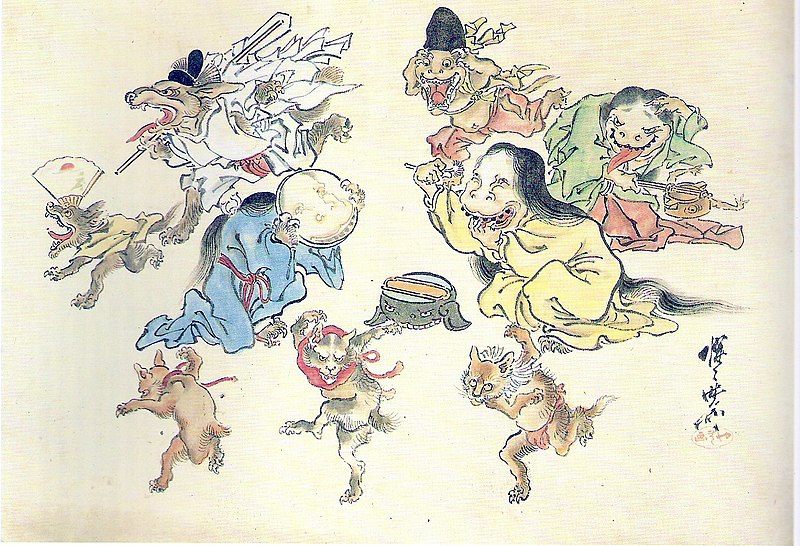}
    \caption{The Night Parade of One Hundred Demons \jp{(百鬼夜行)} by Kawanabe Kyosai. It is said that a parade of supernatural creatures known as yokai march through the streets of Japan at night, and anyone who comes across would be spirited away. The folktale became one of the popular motifs in the Edo period portrayed in many media including ukiyo-e, toys, and picture scrolls.}
    \label{fig:yokai}
\end{figure}

Large Language Models (LLM) have shown remarkable performance in language understanding and generation tasks \cite{NEURIPS2022_b1efde53,touvron2023llama,openai2024gpt4}. 
Despite many LLMs being predominantly trained in English, their generalization capabilities allow them to transfer knowledge across languages, achieving decent performance even in resource-limited languages \cite{chen2023phoenix,shaham2024multilingual,openai2024gpt4}.

While there is evidence for strong cross-lingual transfer capability, cross-cultural transfer is known to be challenging for LLMs \cite{arango-monnar-etal-2022-resources,hershcovich-etal-2022-challenges,lee-etal-2023-hate,huang-yang-2023-culturally,rao2024normad,adilazuarda2024measuring,cao-etal-2024-cultural,liu2024culturally}.
Prior work shows that LLMs tend to be biased toward the values and opinions of certain communities, rather than representing the diversity of human values \cite{santurkar2023opinions,conitzer2024social}.

To address this issue, many studies have investigated methods to evaluate the cultural awareness of LLMs \cite{rao2024normad,adilazuarda2024measuring,liu2024culturally}. Simultaneously, approaches to develop \textbf{culturally aware LLMs} using the language resources of target communities are being explored \cite{pires2023sabia,lin2023taiwan,nguyen2023seallms,huang-etal-2024-acegpt,owen2024komodo,tran2024uccix,etxaniz2024bertaqa}.

While many prior studies have investigated differences in the values and opinions of communities \cite{xu2024exploring,DBLP:conf/aaai/SorensenJHLPWDL24,wang-etal-2024-seaeval,naous2024having,durmus2024measuring}, the \textit{medium} that conveys these values and opinions has received less attention. Specifically, there is less focus on \textbf{folktales}, which are important for creating, receiving, and circulating a community's traditions, values, opinions, and culture \cite{abello2012computational}.

The goal of the study is to investigate how much LLMs know about folktales. Specifically, we focus on \textit{Yokai} (Figure~\ref{fig:yokai}), supernatural phenomena or entities believed to cause such phenomena, which originated from Japanese folktales \cite{KOMATSU2003yokai,foster2024book}.
Yokai have been popular motifs in art and entertainment from the Edo period to the present, appearing in Ukiyo-e (a genre of Japanese paintings), Kabuki (a traditional Japanese theater featuring performance and dance), toys, manga, and anime \cite{kagawa2005yokai,kagawa2006yokai}.

In this paper, we present \textbf{YokaiEval}, a benchmark dataset designed to evaluate knowledge about yokai. YokaiEval is created using Wikipedia articles, with GPT-4o generating the question-answer pairs (QAs). These QAs are then curated using GPT-4omini to filter out inconsistent questions, followed by manual verification to ensure their correctness, appropriateness, and the presence of references. We use YokaiEval to evaluate 31 LLMs, including Japanese-centric, English-centric, and other multilingual models. The results show that Japanese-centric models achieve comparatively higher scores than the others, particularly those models that have been continually pretrained from highly capable English-centric models (e.g., Llama-3). 

This study suggests that using language resources from the community is crucial for acquiring regional folktale knowledge and that continual pretraining is an effective method for training LLMs with these resources.





\section{Yokai: From Traditional Folktales to Today's Art and Entertainment}
\label{sec:yokai}

Yokai are supernatural phenomena or entities believed to cause such phenomena \cite{KOMATSU2003yokai,foster2024book}. Traditionally, yokai are described in oral lore and rumors as \textit{superstitions}, but they have also become popular \textit{characters} in contemporary art and entertainment \cite{kagawa2005yokai,komatsu2009}.


\paragraph{Yokai in folktales.}
Yokai in folktales are thought to have emerged from the human tendency to assign meaning to mysterious phenomena that defy everyday understanding \cite{kagawa2005yokai}. When faced with inexplicable occurrences, humans often experience fear and anxiety. The concept of yokai is believed to have arisen as a means to recognize and thereby alleviate these fears \cite{kagawa2005yokai}.

\paragraph{History of yokai in art and entertainment.}
Although they originate from folktales, yokai have also been recognized as motifs in art and entertainment. By the Edo period, many people in urban areas (e.g., Edo) already viewed the world from a scientific perspective and considered yokai to be fictional rather than real \cite{kagawa2005yokai}. As yokai became recognized as fictional superstitions, people began to enjoy these mysterious characters in the form of art and entertainment.

For example, ghost stories featuring yokai became popular in theater \cite{kagawa2005yokai}. One notable example is Yotsuya Kaidan (\jp{四谷怪談}), a ghost story featuring yokai that remains one of the famous plays in Kabuki today \cite{yotsuyakaidan}. Yokai also became a popular theme in Ukiyo-e. In 1776, Sekien Toriyama painted \textit{The Illustrated Night Parade of a Hundred Demons} (\jp{画図百鬼夜行}), which portrayed each yokai individually, characterizing their unique features \cite{sekien2005}. His art became remarkably popular in the Edo era, and even today, many descriptions of yokai are based on his work \cite{takoshima2018}. Toys featuring yokai were also popular; for instance, \textit{Ghost Playing Cards} (\jp{おばけかるた}) were made in 1860 and gained popularity among children \cite{karuta1998}.

However, rural beliefs and folktales, including those about yokai, were suppressed by the Meiji government in its efforts to Westernize and modernize the country \cite{kagawa2022}. 


\begin{figure*}
    \centering
    \adjustbox{max width=0.85\textwidth}{
    \begin{tikzpicture}[node distance=2cm, auto, align=center]
    \tikzstyle{block} = [rectangle, draw, text centered, minimum height=1cm, minimum width=2.5cm]
    \tikzstyle{dataset} = [cylinder, shape border rotate=90, draw, text centered, aspect=0.5]
    \tikzstyle{arrow} = [thick,->,>=stealth]
    
    \node (base) [block] {Base\\Model};
    \node (pretrain) [block, right of=base, node distance=3.5cm] {Pretraining};
    \node (sft) [block, right of=pretrain, node distance=3.5cm] {Supervised\\Fine-Tuning};
    \node (rlhf) [block, right of=sft, node distance=3.5cm] {Preference\\Learning};
    \node (final) [block, right of=rlhf, node distance=3.5cm] {Final\\LLM};
    
    \draw [arrow] (base) -- (pretrain);
    \draw [arrow] (pretrain) -- (sft);
    \draw [arrow] (sft) -- (rlhf);
    \draw [arrow] (rlhf) -- (final);
    
    \node (doc) [dataset, above of=pretrain, node distance=2cm, minimum height=2.3cm, minimum width=2.3cm] {Cultural\\Document};
    \node (dialogue) [dataset, above of=sft, node distance=2cm, minimum height=1cm, minimum width=1cm] {Cultural\\Dialogue};
    \node (pref) [dataset, above of=rlhf, node distance=2cm, minimum height=1cm, minimum width=1cm] {Cultural\\Preference};
    
    \draw [arrow] (doc) -- (pretrain);
    \draw [arrow] (dialogue) -- (sft);
    \draw [arrow] (pref) -- (rlhf);

\end{tikzpicture}
    }
    \caption{Common approach to training a culturally aware LLM is roughly divided into pretraining, supervised fine-tuning (SFT), and preference learning (PL). One of the questions of the study is to evaluate which step is the most critical to cultural knowledge.}
    \label{fig:training-llm}
\end{figure*}
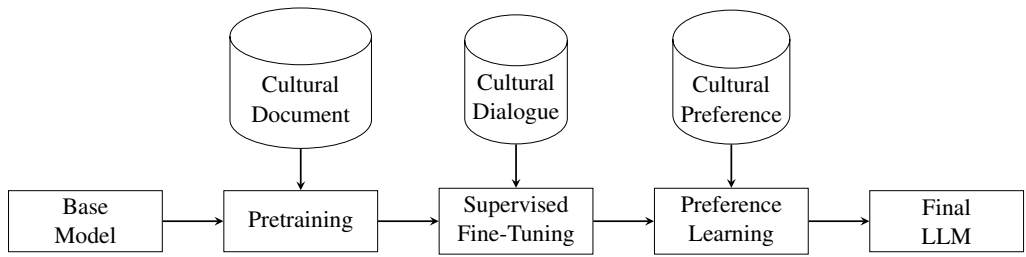

\paragraph{Yokai in today's Japan.}
After the wars, manga artist Shigeru Mizuki revitalized the popularity of yokai through his picture stories and mangas, including GeGeGe no Kitaro \cite{kitaro1986}. Mizuki researched Ukiyo-e and folktales and incorporated these elements into his art \cite{mizuki1974,takoshima2018}, preserving aspects of yokai culture.

Although most people consider yokai to be fictional, stories about them remain present and relevant in today's Japan. For example, the transformation of a human in tragedy into an Oni (a kind of demon in Japanese folktales) is a common plot in these stories. An Ukiyo-e piece from 1902 by Tsukioka Yoshitoshi portrays a woman whose father was killed by a samurai, Omori Hikoshichi, and who then turns into an Oni due to her anger and sorrow \cite{oni1902}. Today, we see similar plots in some of the Oni characters in the series Demon Slayer: Kimetsu no Yaiba \cite{kimetsu2016}.


\paragraph{Summary.}
Yokai is a concept that originated from folktales and has flourished as a popular motif in arts and entertainment, preserved through the efforts of folklorists and artists. They remain popular in today's Japan, both in folktales and in art. In this paper, \textbf{we consider both aspects of yokai to be cultural heritage} that we want LLMs to be knowledgeable about.


\section{Related Work}
We describe prior work on evaluating and training culturally aware LLMs.

\subsection{Evaluating Cultural Awareness of LLM}

While LLMs have shown strong capability in cross-lingual transfer, cross-cultural transfer remains challenging \cite{rao2024normad,adilazuarda2024measuring,liu2024culturally}.
As a result, LLMs often exhibit a bias towards Western culture, where many language resources are more readily available on the internet \cite{santurkar2023opinions,conitzer2024social}. 

Therefore, efforts are being made to evaluate various cultural aspects of LLMs, including values \cite{xu2024exploring,DBLP:conf/aaai/SorensenJHLPWDL24,wang-etal-2024-seaeval}, opinions \cite{naous2024having,durmus2024measuring}, social norms \cite{yu-etal-2024-cmoraleval,rao2024normad,yuan-etal-2024-measuring,agarwal-etal-2024-ethical}, commonsense knowledge \cite{wang-etal-2024-countries,shen-etal-2024-understanding,myung2024blend}, dietary preferences \cite{palta-rudinger-2023-fork,cao-etal-2024-cultural}, and offensive languages \cite{zhou-etal-2023-cultural,lee-etal-2023-hate}. 

Several existing studies have investigated computational approaches to understand folktales \cite{declerck-etal-2012-ontology,lestari-manurung-2015-measuring,declerck-etal-2016-towards,schraagen-2016-folktale,meaney-etal-2024-testing}.
\citet{burda-lassen-2022-ukrainian,burda-lassen-2023-machine} investigate the performance of existing models in translating folktales and highlight the need for a larger folktale corpus with a broader range of cultural terms. \citet{hobson-etal-2024-story} shows that GPT-4o's identification of the values and lessons conveyed in folktales has a high agreement with what human annotators in Western countries identify. \citet{noam2022Folktales,wu-etal-2023-cross} investigate how human values, morals, and gender biases are expressed in folktales across cultures using an international corpus of folktales. 

While many existing studies seek to compare folktales from multiple cultures and compare their differences, we focus on folktales of a single country (Japan).
We argue that establishing a thorough analysis of the folktales within each community is important as it will enhance the quality and depth of multi-cultural comparisons, which we discuss in Appendix~\ref{apd:issue}.

\subsection{Training Culturally Aware LLM}

There are plenty of efforts to train culturally aware LLM.
The training procedures for LLMs can be broadly categorized into three approaches: continual pretraining, supervised fine-tuning, and preference learning.

\paragraph{Continual pretraining.}
Given the huge gap in available language resources between English and other languages, several studies have investigated an approach to train a model starting from an already highly capable English-centric model rather than from scratch.
Continual pretraining has been reported to be effective for learning cultural knowledge across many communities \cite{pires2023sabia,lin2023taiwan,nguyen2023seallms,huang-etal-2024-acegpt,owen2024komodo,tran2024uccix,etxaniz2024bertaqa}. 

\paragraph{Supervised fine-tuning (SFT).}
SFT is a more computationally efficient process than continual pretraining. Many studies have shown that SFT can improve the cultural awareness of LLMs \cite{lin2023taiwan,owen2024komodo,yoo2024hyperclova,huang-etal-2024-acegpt,cahyawijaya2024cendol,zhang2024methodology}. \citet{choenni2024echoes} show that cultural values can be influenced by the language used during the fine-tuning process. On the other hand, \citet{moore-etal-2024-large} indicate that fine-tuned models tend to be more inconsistent on controversial topics.

\paragraph{Preference learning (PL).} 
The effect of preference learning on cultural awareness of LLM has been studied less extensively than pretraining and SFT.
\citet{jinnai-2024-cross} evaluate the effect of intra and cross-cultural alignment, showing that LLMs can learn the commonsense morality of a community with higher accuracy using the dataset annotated by community members.
\citet{chakraborty2024maxminrlhf} propose a method to train a model that reflects the diversity of human preferences.
\citet{yao2025no} show that a naive approach to preference learning can be dominated by the votes of the majority, marginalizing the preference of minorities. 

\begin{figure}[tb]
    \centering
    \includegraphics[width=0.93\columnwidth]{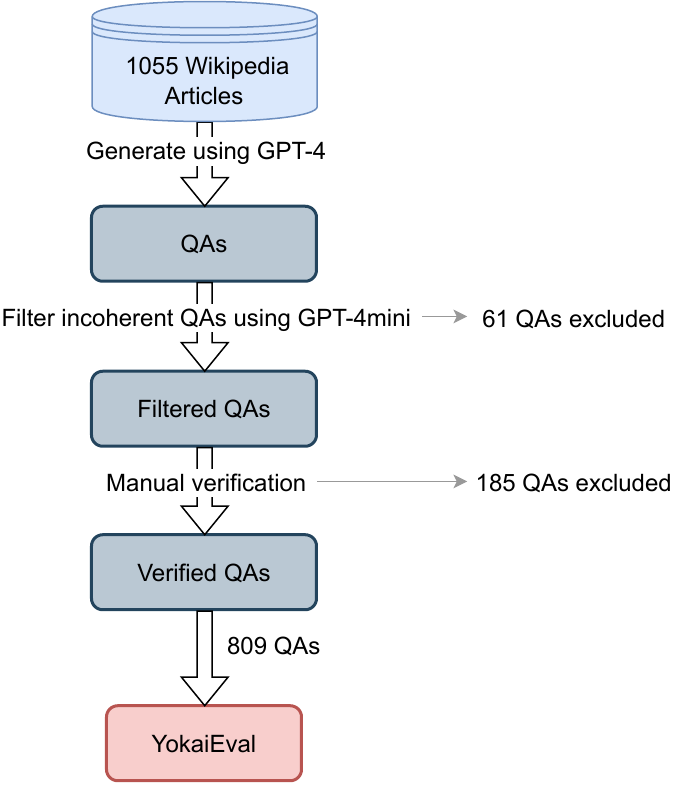}
    \caption{The procedure for generating YokaiEval involves several steps. First, we use articles from Wikipedia as the sources of information and generate QA pairs using GPT-4o. Next, we use GPT-4omini to remove QA pairs that are not coherent with the articles. Finally, the first author manually reviews all the data entries and their references to verify the validity of the QA pairs.}
    \label{fig:dataset-construction}
\end{figure}

\begin{table*}[tb]
    \centering
    \adjustbox{max width=0.95\textwidth}{
    \begin{tabularx}{\textwidth}{XXc}
        \toprule
        Question & Choices & Answer \\
        \midrule
In the case of the Japanese yokai \textit{Oitekebori}, what is the content of the voice that is often heard when trying to leave? & Go back, Leave it, Run away, Go & Leave it\\
When is the day considered particularly important for the appearance of the \textit{Ippondatara}? & January 1, May 5, December 20, August 8 & December 20\\
What is often considered necessary to perceive the existence of the Japanese yokai \textit{Enraenra}? & Having a relaxed mind, Chanting a specific spell, Waiting at a specific place, Watching at a specific time & Having a relaxed mind\\
        \bottomrule
    \end{tabularx}
    }
    \caption{Examples of data entries of the dataset translated to English. The original texts in the dataset are in Japanese.}
    \label{tab:dataset-example}
\end{table*}

\section{Constructing YokaiEval}
\label{sec:dataset}

We first describe the procedure for generating YokaiEval in Section~\ref{sec:construction}. Then, in Section~\ref{sec:dataset-analysis}, we analyze the characteristics of the generated dataset to evaluate if it is suitable for measuring the knowledge of yokai in LLMs.

\subsection{Dataset Construction}
\label{sec:construction}

YokaiEval is constructed using the process described in Figure~\ref{fig:dataset-construction}.

\paragraph{Question generation.}
We generate QAs using 1,054 Wikipedia articles listed in the ``List of Japanese yokai.''\footnote{https://ja.wikipedia.org/wiki/\jp{日本の妖怪一覧}} For each yokai article, we provide the article to GPT-4o and ask it to generate a four-option QA that can be answered using the information within the article. See Appendix~\ref{apd:generation-prompt} for the prompt used for question generation. We manually generate five questions and use them as 5-shot examples.

Since folktales have variations and lack scientific ground truth, we prompt GPT-4o to generate questions that ask which option is often considered true in Japanese folktales, rather than what is scientifically correct.

\paragraph{Automated filtering.}
To ensure that the questions can be solved by someone with knowledge of the yokai, we use GPT-4omini to answer the generated questions using the corresponding articles as prompts (Appendix~\ref{apd:filtering-prompt}). Out of 1,055 QAs, GPT-4omini correctly answers 994 QAs. We manually check some of the QAs that GPT-4omini fails to answer correctly and observe that most of them either lack sufficient clues in the article or are not correctly formed as four-option QAs.

\paragraph{Manual verification.}
For the 994 QAs that GPT-4o answers correctly, we manually check the quality of all the QAs. We evaluate whether each QA is unique, relevant, appropriate, and associated with references. QAs without associated references are excluded to ensure that the questions are verifiable via the references and not based solely on Wikipedia articles written by anonymous contributors. We manually check all the references of the QAs, as we find that GPT-4o often fails to identify the correct references. 

Initially, the first author annotated the data, and two additional annotators subsequently verified these annotations. The verification process ensured that each QA pair met the following criteria:
\begin{enumerate}
    \item The answer to the question is verifiable using the corresponding Wikipedia article.
    \item The statement in the article, which corresponds to the QA and is associated by reference, was written by a domain expert.
    \item The QA contains no inappropriate content.
    \item The QA is not duplicated.
\end{enumerate}
The annotation process was conducted interactively. Whenever an annotator encountered such cases, we held meetings or discussions to decide on the appropriate annotations. Due to this procedure, we do not have a record of inter-annotator agreement.
We exclude 99 QAs for overlapping content, 41 QAs for irrelevant content, 3 for inappropriate content, and 42 QAs for lacking references, resulting in 809 QAs.

Note that the QAs in the dataset do not concern cultural preferences, values, or opinions. The questions assess factual accuracy based on the studies of folklorists and Japanese literary researchers. Each question presents four answer choices, with one correct answer and three incorrect ones. Although folktales themselves exhibit variations, the documented information can be uniquely defined. 

Although the initial data entries come from Wikipedia, the QAs in the resulting dataset (YokaiEval) are all manually verified to be supported by references. Therefore, all QAs are associated with references.

\begin{figure}[tb]
    \centering
    \subfloat[Types of questions in YokaiEval.]{\includegraphics[width=0.95\linewidth]{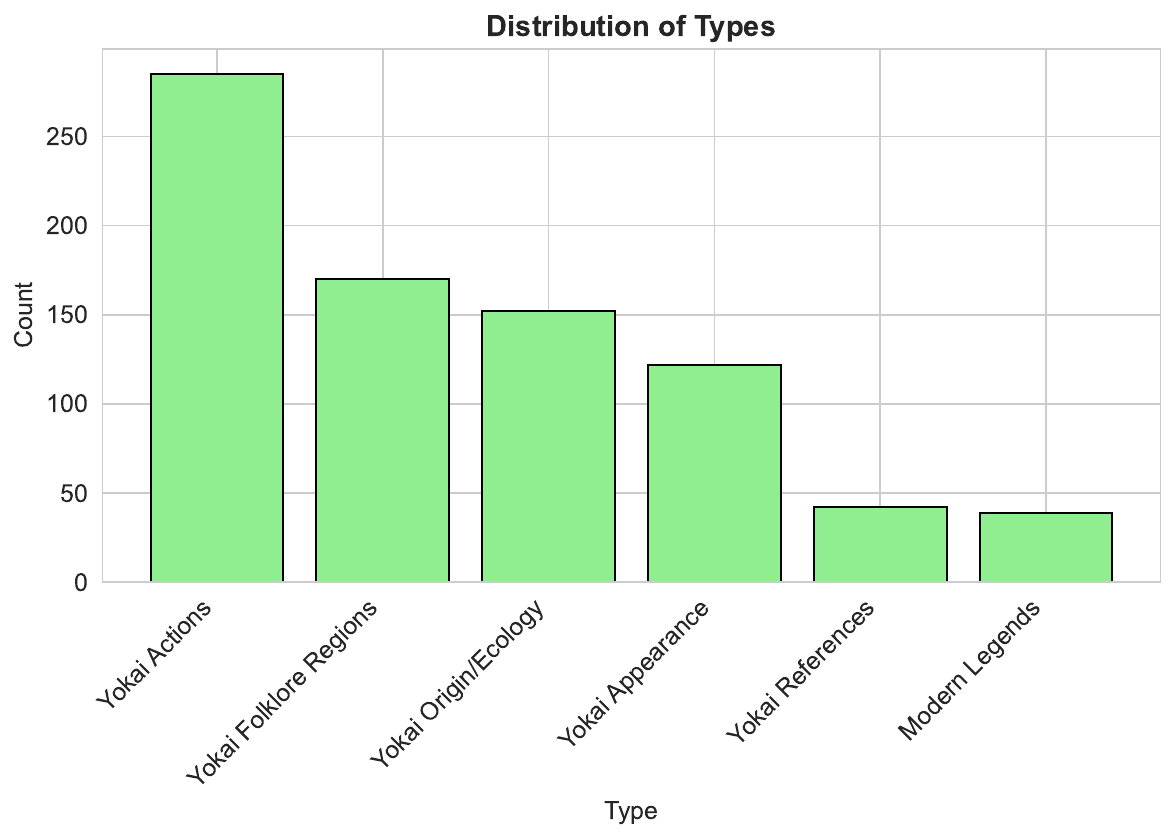}
    \label{fig:question-type}}\\
    \subfloat[Regions of Yokais in YokaiEval.]{\includegraphics[width=0.95\linewidth]{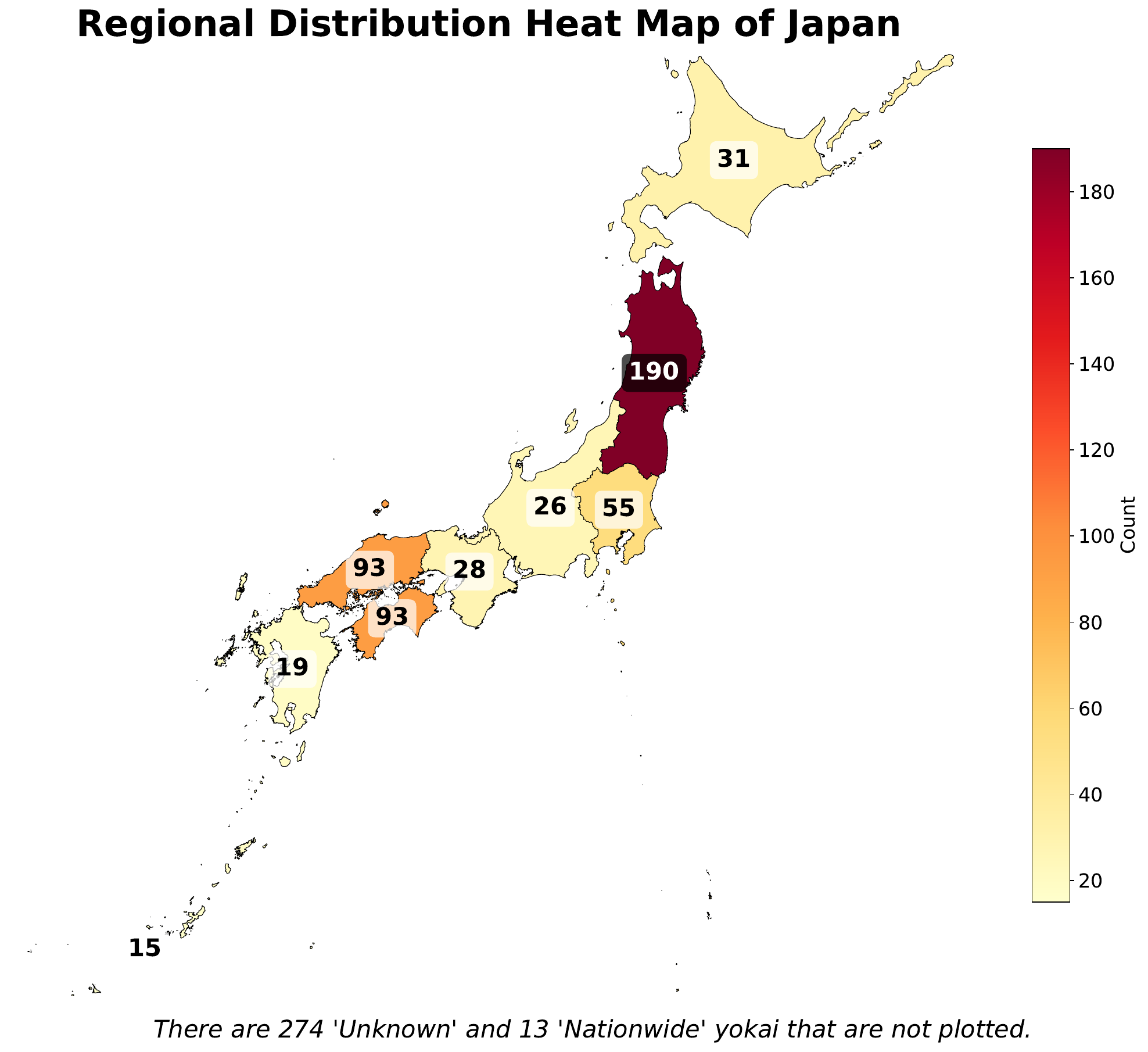}
    \label{fig:region}}    
\end{figure}


\subsection{Analysis of the Generated Dataset}
\label{sec:dataset-analysis}

We analyze the diversity and the quality of the dataset to evaluate if it is valid for measuring knowledge of yokai.

\paragraph{Question types.} 
We classified the question types into six categories using GPT-4o. See Appendix~\ref{apd:analyze-prompt} for the prompt. The number of questions in each category is shown in Figure~\ref{fig:question-type}. The results indicate that answering multiple aspects of yokai is necessary for an LLM to achieve a high score in YokaiEval.

\paragraph{Regions.} 
The region of each yokai is extracted from the corresponding Wikipedia article using GPT-4o (see prompt in Appendix~\ref{apd:analyze-prompt}). Figure~\ref{fig:region} shows the statistics of the regions of the yokai in YokaiEval. Many yokai are difficult to assign to a specific origin because the folktales are spread across multiple regions, resulting in some being labeled as Unknown. Still, the results indicate that the dataset provides good coverage of the regions in Japan.

\begin{table}[tb]
    \centering
    \adjustbox{max width=0.7\columnwidth}{
\begin{tabular}{ll}
\toprule
\textbf{Author} & \textbf{\#refs} \\
\midrule
\jp{村上健司} [Kenji Murakami] & 183 \\
\jp{水木しげる} [Shigeru Mizuki] & 123 \\
\jp{多田克己} [Katsumi Tada] & 127 \\
\jp{柳田國男} [Kunio Yanagita] & 67 \\
\jp{高田衛} [Mamoru Takada] & 65 \\
\jp{鳥山石燕} [Sekien Toriyama] & 60 \\
\jp{稲田篤信} [Atsunobu Inada] & 50 \\
\jp{田中直} [Nao Tanaka] & 50 \\
\jp{草野巧} [Takumi Kusano] & 23 \\
\jp{京極夏彦} [Natsuhiko Kyogoku] & 19 \\
\jp{池田彌三郎} [Yasaburo Ikeda] & 14 \\
\jp{柴田宵曲} [Shokyoku Shibata] & 13 \\
\jp{大藤時彦} [Tokihiko Oto] & 13 \\
\jp{寺島良安} [Ryoan Terashima] & 12 \\
\jp{日野巌} [Iwao Hino] & 11 \\
\jp{藤沢衛彦} [Morihiko Fujisawa] & 11 \\
\jp{千葉幹夫} [Mikio Chiba] & 11 \\
\jp{島田勇雄} [Isao Shimada] & 10 \\
\jp{田野理夫} [Morio Tano] & 10 \\
\jp{戸部民夫} [Tamio Tobe] & 10 \\
\bottomrule
\end{tabular}
}
    \caption{The list of authors with more than or equal to ten references to their work in the QAs of YokaiEval.}
    \label{tab:authors}
\end{table}

\paragraph{References.}
Table~\ref{tab:authors} shows the authors whose work is cited ten or more times in YokaiEval. We observe a good balance among contemporary artists (e.g., Shigeru Mizuki and Natsuhiko Kyogoku), artists from the Middle Ages (e.g., Sekien Toriyama), and scholars of folklore and Japanese literature (e.g., Kunio Yanagita, Atsunobu Inada, Yasaburo Ikeda, Tokihiko Oto, and Iwao Hino). A list of frequently referenced materials is provided in Appendix~\ref{apd:reference}.

\paragraph{Questions solvable with reasoning.}
We observe that some questions can be guessed if one knows the tendencies of Japanese folktales and the meanings of the names of the yokai. For example, \textit{Oitekebori} (\jp{置行堀}; Table~\ref{tab:dataset-example}) literally means "left behind in the ditch." Thus, one can guess that "Leave it" is likely to be the correct answer. On the other hand, some questions are difficult to guess without specific knowledge of the particular yokai. For example, it is extremely difficult to infer that \textit{Ippondatara} (\jp{一本だたら}; Table~\ref{tab:dataset-example}) appears on December 20th without specific knowledge of this yokai, making it extremely difficult to deduce from the text alone.

Therefore, while some questions are solvable with inductive knowledge of Japanese folktales, others require specific knowledge of the particular yokai. We consider this to be a desirable feature of the dataset, as our goal is to evaluate general knowledge of yokai.

\paragraph{Offensive and biased contents.}
By nature, some folktales contain stories or views that can be harmful (e.g., Kappa in \citealp{tohno1976}). We find that most of the QAs generated by GPT-4o do not include such content. Only three QAs were excluded during manual verification due to potentially inappropriate content. We hypothesize that this is due to using GPT-4o via the Azure OpenAI API. GPT-4o is aligned to avoid generating potentially harmful content, and the Azure OpenAI API includes a content moderation filter that prevents the generation of harmful content.

Although offensive and biased content in folktales is also important for folktale studies, our research questions do not necessarily require the inclusion of such content for assessment.

\begin{figure*}[t]
    \centering
    \includegraphics[width=0.9\textwidth]{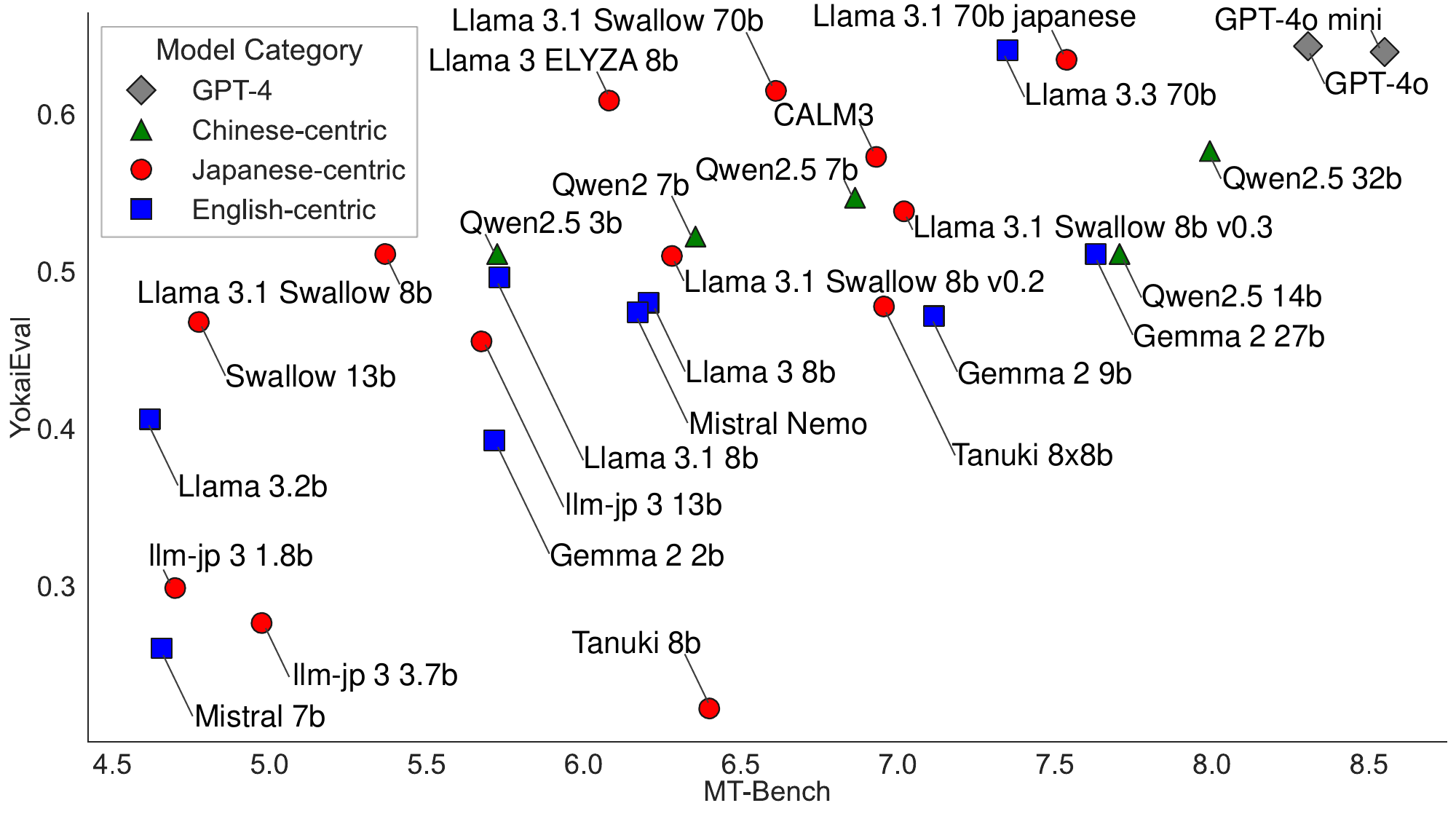} 
    \caption{Scores of YokaiEval and JMT-Bench. Red: Japanese-centric models (including models continually pretrained on Japanese), Blue: English-centric languages models, Green: Chinese-centric models. Gray: GPT-4o models which we use for the dataset construction, which may have biases. The scores are included as references. See Appendix~\ref{apd:raw-score} for the exact values of the scores.}
    \label{fig:mt-jf-bench}
\end{figure*}

\section{Evaluating Japanese and Multilingual LLMs on YokaiEval}
\label{sec:evaluation}

With YokaiEval generated, we run experiments to evaluate the performance of 31 publically available LLMs, including Japanese-centric, English-centric, and Chinese-centric models.
We then further look into the performance of Japanse-centric models to investigate which training procedure (pretraining, SFT, and PL) is important to acquiring the knowledge of yokai.

\subsection{Experimental Setup}

\begin{figure}[tb]
    \centering
\begin{CJK}{UTF8}{ipxm}
\begin{Verbatim}[breaklines,frame=single,breaksymbolleft=]
以下に、日本の妖怪に関する質問をする指示があります。質問に対する回答を記述してください。(Translation: Below are instructions for asking questions about Japanese yokai. Please provide answers to the questions.)

{question}
- {option1}
- {option2}
- {option3}
- {option4}
\end{Verbatim}
\end{CJK}
    \caption{The prompt for generating the response from LLMs for YokaiEval.}
    \label{tab:prompt}
\end{figure}

We use the prompt in Table~\ref{tab:prompt} to generate outputs from the LLMs. See Appendix~\ref{apd:hyperparams} for the generation hyperparameters. We use GPT-4o to determine if the responses are correct, rather than relying on lexical matching, because the notation of yokai inherently has a lot of fluctuations. See Appendix~\ref{apd:eval-prompt} for the prompt used for evaluation. As far as we are aware, there were no errors in the GPT-4o evaluation.\footnote{We implemented a lexical matching parser capable of extracting answers from approximately 80\% of the LLMs. For those answers extracted through lexical matching, the same answers were identified by GPT-4o. Additionally, we manually reviewed the evaluation results of GPT-4o provided by two LLMs and found no errors.}

We use Japanese MT-Bench (JMT-Bench) as a reference for the generic capability of the LLMs in Japanese.\footnote{\url{https://github.com/Stability-AI/FastChat/tree/jp-stable}}
See Appendix~\ref{apd:jmt-score} for details.

\subsection{Results}
\label{sec:results}

Figure~\ref{fig:mt-jf-bench} shows the scores on YokaiEval and JMT-Bench. The YokaiEval score is computed as the number of correct answers divided by the total number of questions. Overall, Japanese-centric models achieve higher scores on YokaiEval compared to models with similar performance on the JMT-Bench. Notably, models that undergo continual pretraining from the Llama-3 family tend to perform particularly well on YokaiEval.

The study suggests that training LLMs with Japanese language resources improves performance on YokaiEval, suggesting a greater knowledge of Japanese folktales. In particular, continually pretraining a highly capable English-centric model (e.g., the Llama-3 family) on Japanese documents appears to be important for acquiring knowledge of Japanese folktales.

\paragraph{The effect of SFT and PL.}

\begin{table}
    \centering
    \begin{tabular}{cc}
    \toprule
        Model & YokaiEval \\
    \midrule
        (meta-llama-3.1-8b-instruct) & 0.496 \\
        llama-3.1-swallow-8b-instruct-v0.1 & 0.511 \\
        llama-3.1-swallow-8b-instruct-v0.2 & 0.510 \\
        llama-3.1-swallow-8b-instruct-v0.3 & 0.538 \\
    \bottomrule
    \end{tabular}
    \caption{The effect of SFT on the YokaiEval scores.}
    \label{tab:sft}
\end{table}

\begin{table*}
    \centering
    \begin{tabular}{cccc}
    \toprule
         & SFT model & DPO Model & Improvement \\
    \midrule
        calm2-7b-chat & 0.242  & 0.199 & -0.043 \\
        llm-jp-13b-instruct-full-jaster-v1.0 & 0.154 & 0.119 & -0.036 \\
        swallow-7b-instruct-v0.1 & 0.340 & 0.424 & 0.084 \\
    \bottomrule
    \end{tabular}
    \caption{The effect of DPO on the YokaiEval scores.}
    \label{tab:dpo}
\end{table*}

As shown in Figure~\ref{fig:mt-jf-bench}, models continually pretrained on Japanese datasets achieve high scores on YokaiEval. These models also undergo SFT and PL in addition to continual pretraining. This raises the question of whether fine-tuning alone is sufficient for acquiring cultural knowledge. To investigate this, we conduct an observational ablation study to assess whether SFT and PL alone effectively improve YokaiEval scores. Note that our focus is on the effects of standard fine-tuning processes rather than fine-tuning specifically designed to enhance knowledge of yokai.

Table~\ref{tab:sft} presents the YokaiEval scores for the llama-3.1-swallow-8b-instruct series. These models are continually pretrained on llama-3.1 and llama-3.1-instruct. Additional SFT processes are applied between versions v0.1, v0.2, and v0.3. While SFT leads to a significant improvement in JMT-Bench scores, its impact on YokaiEval is marginal, suggesting that SFT may have a limited effect on acquiring knowledge of Japanese folktales.

Since only a few models are available both before and after the PL process, we trained models ourselves to evaluate the effect of PL. Specifically, we used ChatbotArena-Ja \cite{jinnai-2024-cross} to train three LLMs with direct preference optimization (DPO) \cite{rafailov2023direct} using LoRA \cite{hu2022lora}.

ChatbotArena-Ja is a dataset consisting of English instructions from Chatbot Arena \cite{NEURIPS2023_91f18a12}, translated into Japanese, along with responses generated by Japanese LLMs. This dataset has been shown to improve Japanese LLM performance on the JMT-Bench. Since the instructions originate from English sources, it is unlikely to contain knowledge relevant to yokai. By using this dataset, we can isolate the effect of the PL procedure itself, independent of any additional knowledge introduced through training data. See Appendix~\ref{apd:hyperparams} for the hyperparameters used in the DPO process.

Table~\ref{tab:dpo} presents the scores of models before and after DPO. Overall, the improvements are inconsistent across models, suggesting that DPO on ChatbotArena-Ja may not be effective in improving performance on YokaiEval.

Note that these results do not imply that SFT and PL on a dataset specifically designed for yokai knowledge would fail to improve YokaiEval scores. Rather, our findings indicate that fine-tuning on a standard dataset does not significantly impact YokaiEval performance. 
The purpose of the experiment is to separate out the effect of SFT and PL so that we can evaluate the effect of continual pertaining more precisely.

The result shows that the post-training processes are less likely to be effective for improving performance on YokaiEval compared to JMT-Bench, which implies that continual pertaining is likely to be the key step in acquiring knowledge of Japanese folktales.

\begin{figure}
    \centering
    \includegraphics[width=0.95\columnwidth]{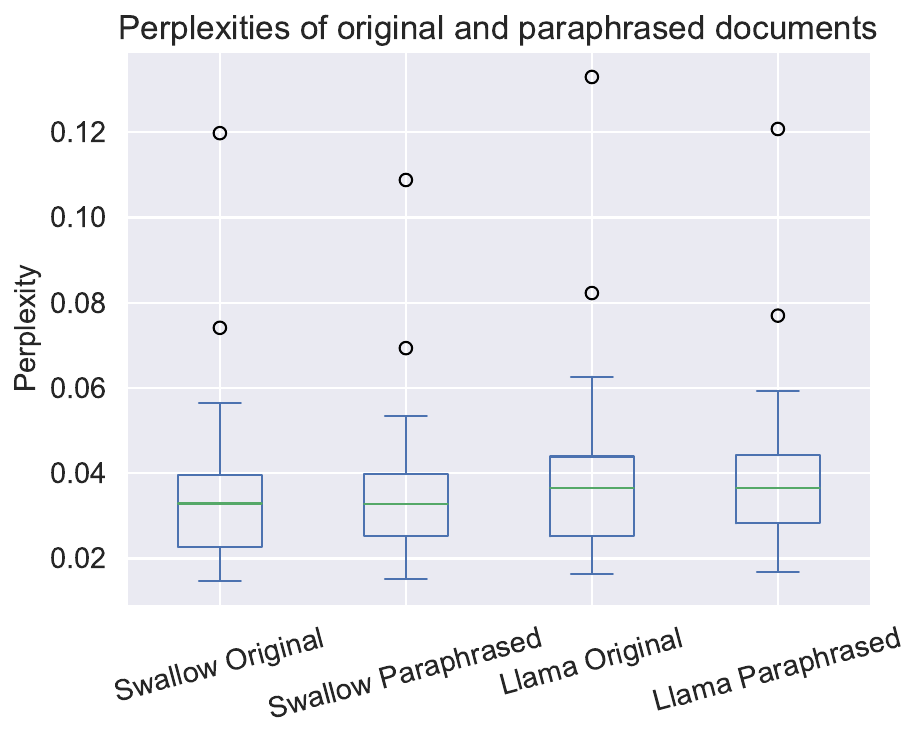}
    \caption{The distribution of perplexities of the original and paraphrased articles to assess the potential of data leakage.}
    \label{fig:leakage}
\end{figure}

\begin{table}[tbh]
    \centering
    \adjustbox{max width=\columnwidth}{
\begin{tabular}{lrrr}
\toprule
\textbf{Types of questions} & \textbf{\#Questions} & \textbf{Accuracy} \\
\midrule
Regions of Yokai & 170 & 59.41\% \\
Yokai's References & 42 & 69.05\% \\
Yokai's Appearance & 122 & 66.39\% \\
Yokai's Origin and Ecology & 153 & 54.25\% \\
Actions of Yokai & 285 & 64.21\% \\
Modern Legends & 38 & 55.26\% \\
\bottomrule
\end{tabular}
}
    \caption{Accuracy of Swallow-70b in YokaiEval for each question type.}
    \label{tab:error-type}
\end{table}

\begin{table}[tbh]
    \centering
    \begin{tabular}{lrr}
    \toprule
    \textbf{Regions of Yokai} & \textbf{\#Questions} & \textbf{Accuracy} \\
    \midrule
    Unknown & 274 & 67.88\% \\
    Chugoku / Shikoku & 93 & 54.84\% \\
    Chubu & 26 & 61.54\% \\
    Kyushu & 19 & 78.95\% \\
    Hokkaido & 31 & 58.06\% \\
    Nationwide (Japan) & 13 & 61.54\% \\
    Tohoku & 190 & 60.00\% \\
    Okinawa & 15 & 40.00\% \\
    Kinki & 28 & 64.29\% \\
    Kanto & 55 & 56.36\% \\
    \bottomrule
    \end{tabular}
    \caption{Accuracy of Swallow-70b in YokaiEval for each region of Yokai.}
    \label{tab:error-region}
\end{table}

\paragraph{Analysis of potential data leakage.}

A potential issue with using Wikipedia articles is that they may have been included in the training data of the LLMs, raising concerns about the suitability of YokaiEval for evaluation.

To investigate this, we follow the method of \citet{deng2024benchmark} by generating paraphrased versions of 20 randomly selected articles using GPT-4o and comparing the perplexity of the LLMs. If an LLM has ``memorized'' an article, it should exhibit significantly lower perplexity for the original article compared to its paraphrased version.

We evaluate Swallow (\texttt{llama\--3.1\--swallow\-8b\--instruct\--v0.1}) and Llama (\texttt{meta\--llama\-3.1\--8b\--instruct}), as these models and their derivatives achieve high scores on YokaiEval.
Figure~\ref{fig:leakage} shows the perplexity of both models on the original and paraphrased articles. Overall, no significant difference is observed between the two groups of articles for either model. This suggests no clear evidence of data leakage in either case. We speculate that articles about yokai are often overlooked by developers as they are unlikely to be relevant for existing benchmark datasets.

\paragraph{Error analysis.}

Table~\ref{tab:error-type} shows the accuracy of the Swallow-70b (\texttt{llama-3.1-swallow-70b-instruct-v0.1}) model on each type of question in YokaiEval. Overall, the accuracy is balanced over the types. 
Table~\ref{tab:error-region} shows the accuracy of the model on each region of Yokai. Overall, the accuracy is balanced except for Okinawa, where the accuracy is only 40\%. It may be because Okinawa has a historically distinct background compared to the other regions, as it was once the Ryukyu Kingdom and the United States. In-depth analysis is required to analyze the effect of the regions to the knowledge of the LLMs.

\section{Conclusions}
\label{sec:conclusions}
In this paper, we introduce YokaiEval, a benchmark dataset designed to evaluate LLMs' knowledge of Japanese folktales. The dataset is generated by extracting QAs from Wikipedia articles using GPT-4o, followed by curation with GPT-4omini and human verification.

We evaluate 31 LLMs using YokaiEval, showing that Japanese LLMs outperform the others, with models continually pretrained on Japanese language resources achieving the highest scores. The result implies the importance of using community-sourced resources to develop LLMs knowledgeable of folktales.


\section{Limitations}

Our study focuses on Yokai in Japanese folktales, and the findings may differ when applied to other folktales, communities, or cultural domains. Our methodology may be extended to other cultural contexts in future research.

The dataset used in this study only includes yokai recorded in digital media (i.e., Wikipedia) up to the present day. Thus, it may be biased and contain underrepresented regions or eras. It may not represent the folktales in their original forms. Additionally, the QAs were extracted using GPT-4o, which may introduce bias into the dataset.

YokaiEval is not intended to serve as a standalone benchmark for assessing the general cultural understanding of LLMs. Instead, we expect it to be most useful when combined with other benchmarks (e.g., Japanese MT-Bench). Since YokaiEval addresses an area not covered by existing benchmarks, it serves as a valuable complementary evaluation dataset.

In 2025, the National Diet Library of Japan holds approximately 5,590 digital records, 10,800 books and articles not available in digital form, 2,313 research articles, and 2,130 multimedia records related to Yokai.\footnote{\url{https://www.ndl.go.jp/}} Using non-digital resources to further enhance the dataset is future work.

The analysis in Section~\ref{sec:evaluation} is observational, relying on publicly available models rather than models trained specifically for this analysis. Therefore, several factors, beyond the cultural training procedures we discuss, could influence the YokaiEval scores. For instance, some Japanese LLMs use tokenizers specifically designed for Japanese, which may contribute to their higher performance on YokaiEval \cite{fujii2024continual,aizawa2024llm}.

Our analysis is limited to evaluating the knowledge embedded in the LLMs themselves. However, it may not be the only way to achieve culturally aware LLMs. One potential approach for acquiring cultural knowledge is Retrieval-Augmented Generation (RAG), which generates output by using a search engine or vector database to retrieve relevant information during inference \cite{lewis2020retrieval}.
 





\section{Ethical Considerations}

The dataset is constructed from publicly available Wikipedia articles. 
Analysis of folktales should be done carefully, as it may promote biased or polarized views of the community.
We encourge users of YokaiEval to be mindful of the potential harm when using the dataset.


\section*{Acknowledgments}
We thank the anonymous reviewers for their insightful comments on the manuscript.
We would like to thank our colleagues at CyberAgent AI Lab for the helpful discussions.


\jp{
\bibliography{anthology,ms,ms2,msculture,yokai}
}

\appendix

\section{Prompts for Constructing YokaiEval  (Section~\ref{sec:construction})}
YokaiEval is constructed using GPT-4o in various ways.
This Appendix lists the prompts used for each of the processes on the dataset construction.

\subsection{Prompt for Generating QAs}
\label{apd:generation-prompt}

Figure~\ref{fig:qa_generation_prompt} shows the 5-shot prompt used to generate a QA from a Wikipedia article using GPT-4o. 

\begin{figure*}
    {\scriptsize
\begin{CJK}{UTF8}{ipxm}
\begin{Verbatim}[breaklines,frame=single,breaksymbolleft=]
###Instruction###
日本の妖怪{yokai_name}の知識を確認する質問を1つ作成しなさい。
###conditions###
1. 説明文から検証可能な質問にしなさい
2. 4択問形式の質問にしなさい
3. その妖怪についての知識がないと正解出来ない質問にしなさい
4. 妖怪の名前から推測不可能な質問にしなさい
5. 妖怪に関する記述は文献によって異なる場合が多々あるため、「とされていることが多いでしょうか？」のように質問し、回答の存在を保証できるような質問にしなさい
6. コードブロックのないjson形式にしなさい
7. 必ず予想では正解出来ない質問を作りなさい
###{yokai_name}についての説明文###
{detail}
###Example###

{{
    "quesiton": "日本の妖怪である「赤えい」の大きさはどの程度とされていることが多いでしょうか？以下の4つから回答を1つ選び出力しなさい。",
    "choices": [
        "10メートル",
        "100メートル",
        "1キロメートル",
        "10キロメートル"
    ],
    "answer": "10キロメートル"
}}

{{
    "quesiton": "日本の妖怪である「鍛冶媼」の正体とされている動物はなんとされていることが多いでしょうか？以下の4つから回答を1つ選び出力しなさい。",
    "choices": [
        "牛",
        "鼠",
        "狼",
        "馬"
    ],
    "answer": "狼"
}}

{{
    "question": "日本の妖怪である「片耳豚」にあることをされると人間は魂を抜かれてしまうとされています。何をされると魂を抜かれてしまうとされていることが多いでしょうか？以下の4つから回答を1つ選び出力しなさい。",
    "choices": [
        "影を踏まれる",
        "股の下をくぐられる",
        "見つめられる",
        "前を横切る"
    ],
    "answer": "股の下をくぐられる"
}}

{{
    "question": "日本の妖怪である「木心坊」はある木を材料にすりこぎを作るとその木が変化して生まれるといわれています。何の木を材料にすると生まれるとされていることが多いでしょうか？以下の4つから回答を1つ選び出力しなさい。",
    "choices": [
        "桜",
        "椿",
        "樅",
        "欅"
    ],
    "answer": "椿"
}}

{{
    "question": "日本の妖怪「馬魔」は馬に何かするとされています。何をするとされていることが多いでしょうか？以下の4つから回答を1つ選び出力しなさい。",
    "choices": [
        "転ばす",
        "殺す",
        "乗る",
        "吹き飛ばす"
    ],
    "answer": "殺す"
}}
\end{Verbatim}
\end{CJK}
}
    \caption{Prompt for generating a QA from a Wikipedia article using GPT-4o. \textbf{The English translation is available in Figure~\ref{fig:qa_generation_prompt-en}.}}
    \label{fig:qa_generation_prompt}
\end{figure*}

\begin{figure*}
    {\scriptsize
\begin{Verbatim}[breaklines,frame=single,breaksymbolleft=]
###Instruction###
Create a question to test knowledge about the Japanese yokai {yokai_name}.
###Conditions###
1. Make the question verifiable from the description.
2. Make the question a multiple-choice question with four options.
3. Make the question one that cannot be answered correctly without knowledge of the yokai.
4. Make the question one that cannot be guessed from the yokai's name.
5. Since descriptions of yokai often vary by source, frame the question in a way that ensures the existence of an answer, such as "What is often said to be...?"
6. Format the question in JSON without any code blocks.
7. Ensure the question cannot be answered correctly by guessing.
###Example###
{{
    "question": "What is often said to be the size of the Japanese yokai 'Akaei'? Choose one answer from the following four options.",
    "choices": [
        "10 meters",
        "100 meters",
        "1 kilometer",
        "10 kilometers"
    ],
    "answer": "10 kilometers"
}}
{{
    "question": "What animal is often said to be the true form of the Japanese yokai 'Kajibokuro'? Choose one answer from the following four options.",
    "choices": [
        "Cow",
        "Rat",
        "Wolf",
        "Horse"
    ],
    "answer": "Wolf"
}}
{{
    "question": "What action is often said to cause a human to lose their soul when performed by the Japanese yokai 'Kataeributa'? Choose one answer from the following four options.",
    "choices": [
        "Stepping on their shadow",
        "Passing under their legs",
        "Staring at them",
        "Crossing in front of them"
    ],
    "answer": "Passing under their legs"
}}
{{
    "question": "What type of tree is often said to give birth to the Japanese yokai 'Kishinbo' when used to make a pestle? Choose one answer from the following four options.",
    "choices": [
        "Cherry",
        "Camellia",
        "Fir",
        "Zelkova"
    ],
    "answer": "Camellia"
}}
{{
    "question": "What is often said to happen to horses when the Japanese yokai 'Umama' is around? Choose one answer from the following four options.",
    "choices": [
        "They are tripped",
        "They are killed",
        "They are ridden",
        "They are blown away"
    ],
    "answer": "They are killed"
}}        
\end{Verbatim}
    }
    \caption{Prompt for generating a QA from a Wikipedia article using GPT-4o (Figure~\ref{fig:qa_generation_prompt}) translated to English.}
    \label{fig:qa_generation_prompt-en}
\end{figure*}

\subsection{Prompt for Filtering QAs}
\label{apd:filtering-prompt}

Figure \ref{fig:correctness_prompt} shows the prompt used for the filtering process. The filtering process tests if the QA is generated in a way that can be solved given the information of the corresponding article.

\begin{figure*}
    \centering
    {\footnotesize
\begin{CJK}{UTF8}{ipxm}
\begin{Verbatim}[breaklines,frame=single,breaksymbolleft=]
###Task###
これは{yokai_name}の説明です。説明を参考にして、質問に答えてください。
###説明###
{yokai_detail}。
###質問###
{question}
\end{Verbatim}
\end{CJK}
}
    \caption{Prompt for evaluating the correctness of QAs using GPT-4o. \textbf{The English translation is available at Figure~\ref{fig:correctness_prompt-en}.}}
    \label{fig:correctness_prompt}
\end{figure*}

\begin{figure*}
    {\footnotesize
\begin{Verbatim}[breaklines,frame=single,breaksymbolleft=]
###Task###
This is a description of {yokai_name}. Please refer to the description and answer the question.
###Description###
{yokai_detail}.
###Question###
{question}
\end{Verbatim}
}
    \caption{Prompt for evaluating the correctness of QAs using GPT-4o (Figure~\ref{fig:correctness_prompt}), translated to English.}
    \label{fig:correctness_prompt-en}
\end{figure*}

\subsection{Prompt for Extracting the Reference of the QA}
\label{apd:reference-prompt}

Figure~\ref{fig:reference_prompt} is the prompt we use to identify the reference that corresponds to the QA using GPT-4o (Section~\ref{sec:construction}). We find that many of the references identified by GPT-4o are incorrectly attributed, so we correct them manually in the manual verification process.

\begin{figure*}
    {\footnotesize
\begin{CJK}{UTF8}{ipxm}
\begin{Verbatim}[breaklines,frame=single,breaksymbolleft=]
[[Task]]
以下に示す質問は[detail]の情報を元に作成されました。
質問と解答が基づいている情報源・文献を、detailの中の番号で出力してください。
複数存在する場合は複数出力してください。例えば、[番号]
質問が文献に基づいていない場合は[0]と出力してください。
[[質問]]
{question}
{choices}
[[解答]]
{answer}
[[detail]]
{yokai_name}
{yokai_detail}
\end{Verbatim}
\end{CJK}
}
    \caption{Prompt for identifying the reference of the QA using GPT-4o. \textbf{The English translation is available in Figure~\ref{fig:reference_prompt-en}.}}
    \label{fig:reference_prompt}
\end{figure*}

\begin{figure*}
    {\footnotesize
\begin{Verbatim}[breaklines,frame=single,breaksymbolleft=]
[[Task]]
The following question was created based on the information in [detail].
Please output the source(s) or document(s) that the question and answer are based on, using the number(s) in the detail. If there are multiple sources, output multiple numbers, for example, [number]. If the question is not based on any document, output [0].
[[Question]]
{question}
{choices}
[[Answer]]
{answer}
[[Detail]]
{yokai_name}
{yokai_detail}
\end{Verbatim}
}
    \caption{Prompt for identifying the reference of the QA using GPT-4o (Figure~\ref{fig:reference_prompt}), translated into English.}
    \label{fig:reference_prompt-en}
\end{figure*}

\begin{figure*}
    \centering
    \includegraphics[width=0.7\textwidth]{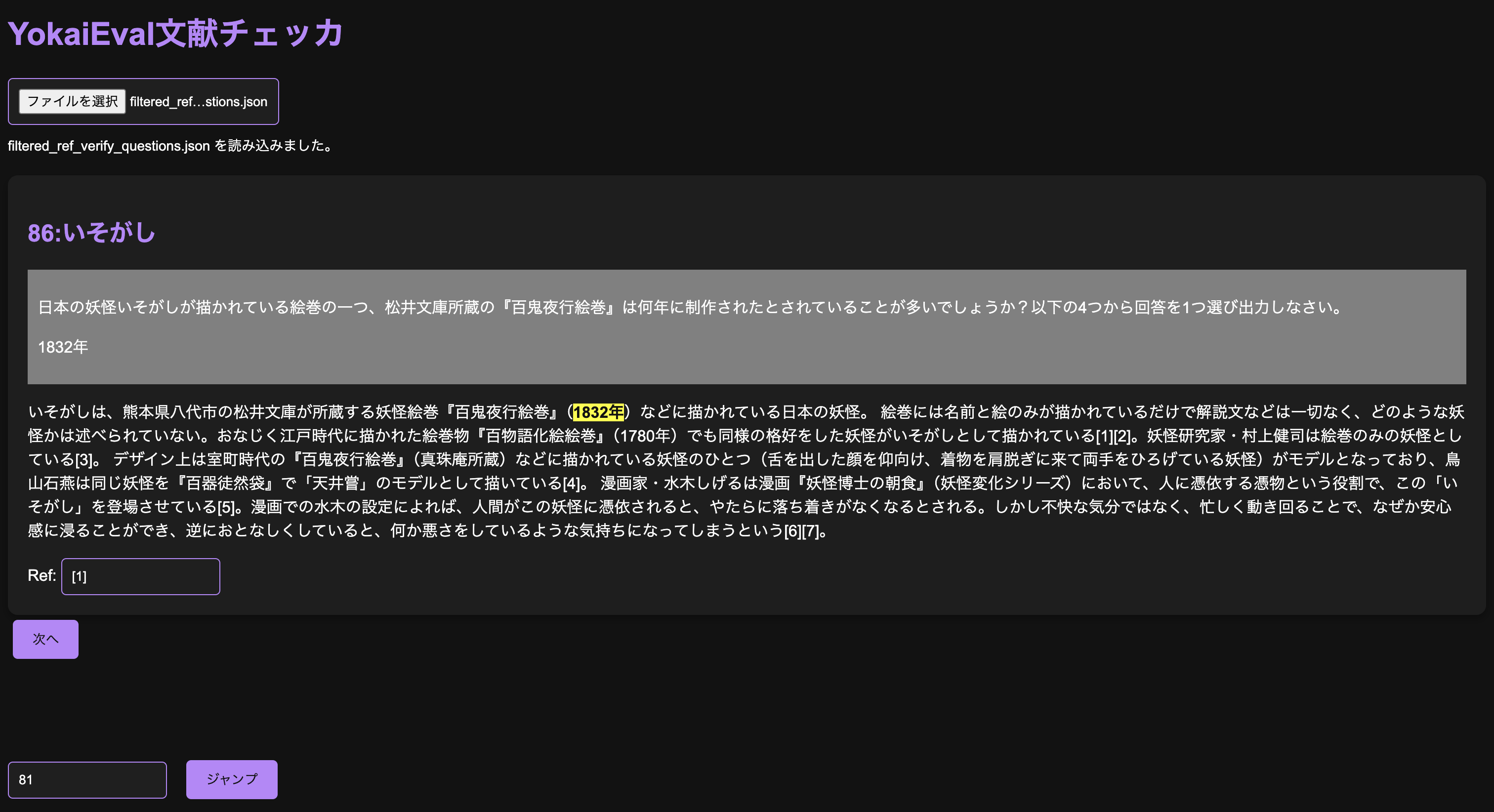}
    \caption{Snapshot of the annotation tool developed for the manual verification step.}
    \label{fig:annotation-tool}
\end{figure*}

\section{Annotation Tool for Manual Verification}
\label{sec:annotation-tool}

We develop a lightweight annotation tool to manually verify the data entries.
Figure \ref{fig:annotation-tool} is the snapshot of the annotation tool. The tool is designed to show the Wikipedia article, the generated QA, and the question reference. The task is to manually check if the QA is appropriate and has a reference. 
The interface is designed to edit the reference as it has the highest frequency or errors.

\begin{figure*}
    {\footnotesize
\begin{CJK}{UTF8}{ipxm}
\begin{Verbatim}[breaklines,frame=single,breaksymbolleft=]
以下は妖怪に関する質問と回答です。この質問は以下のどの質問に分類されますか？
[[回答]]という形式で答えてください。例えば [[A]]と答えてください。
これら以外の分類である場合は自由記述で答えてください。
A. 妖怪の行う行動
B. 妖怪の形状・見た目
C. 妖怪の生まれ・生態
D. 妖怪の典拠
E. 妖怪の伝承のある地域
F. 現代における伝承
{question}
\end{Verbatim}
\end{CJK}
}
    \caption{Prompt for categorizing the type of the QA using GPT-4o. \textbf{The English translation is available in Figure~\ref{fig:question-type-prompt-en}.}}
    \label{fig:question-type-prompt}
\end{figure*}

\begin{figure*}
    {\footnotesize
\begin{Verbatim}[breaklines,frame=single,breaksymbolleft=]
The following is a question and answer about a yokai. Which category does this question fall into? Please answer in the format [[Answer]]. For example, answer [[A]]. If it falls into a category other than those listed, please answer in free text.
A. Actions performed by the yokai
B. Appearance of the yokai
C. Origin or ecology of the yokai
D. References to the yokai
E. Region where the yokai is traditionally known
F. Modern-day traditions
{question}
\end{Verbatim}
}
    \caption{Prompt for categorizing the type of the QA using GPT-4o (Figure~\ref{fig:question-type-prompt}), translated to English.}
    \label{fig:question-type-prompt-en}
\end{figure*}

\begin{figure*}
    {\footnotesize
\begin{CJK}{UTF8}{ipxm}
\begin{Verbatim}[breaklines,frame=single,breaksymbolleft=]
以下は日本の妖怪「{yokai_name}」に関する説明です。{yokai_name}が伝わる地域はどこであると記述されていますか？
[[回答]]という形式で答えてください。例えば [[A]]と答えてください。
日本全国でな無く、かつ複数にまたがる場合は自由記述で答えてください。
これら以外の分類である場合は自由記述で答えてください。
A. 日本全国
B. 北海道
C. 東北
D. 関東
E. 中部地方
F. 近畿地方
G. 中国地方
H. 四国地方
I. 九州
J. 沖縄
K. 不明
{article}
\end{Verbatim}
\end{CJK}
}
    \caption{Prompt for extracting the region of the yokai from the Wikipedia article using GPT-4o. \textbf{The English translation is available in Figure~\ref{fig:region-prompt-en}.}}
    \label{fig:region-prompt}
\end{figure*}

\begin{figure*}
    {\footnotesize
\begin{Verbatim}[breaklines,frame=single,breaksymbolleft=]
The following is an explanation about the Japanese yokai "{yokai_name}". Where is the region mentioned where {yokai_name} is traditionally known?
Please answer in the format [[Answer]]. For example, answer [[A]].
If it is not nationwide in Japan and spans multiple regions, please answer in free text.
If it falls into a category other than those listed, please answer in free text.
A. Nationwide in Japan
B. Hokkaido
C. Tohoku
D. Kanto
E. Chubu region
F. Kinki region
G. Chugoku region
H. Shikoku region
I. Kyushu
J. Okinawa
K. Unknown
{article}
\end{Verbatim}
}
    \caption{Prompt for extracting the region of the yokai from the Wikipedia article using GPT-4o (Figure~\ref{fig:region-prompt}), translated to English.}
    \label{fig:region-prompt-en}
\end{figure*}

\section{Prompts for Analyzing YokaiEval (Section~\ref{sec:dataset-analysis})}
\label{apd:analyze-prompt}

Figure~\ref{fig:question-type-prompt} shows the prompt used to categorize the type of question.
Figure~\ref{fig:region-prompt} shows the prompt used to extract the region of the yokai from the Wikipedia article.

\section{List of Frequently Referred Materials}
\label{apd:reference}
Table~\ref{tab:reference} is a list of frequently referred materials used in the QAs of YokaiEval.

\begin{table*}
    \centering
\begin{tabularx}{\linewidth}{Xc}
    \toprule
Book & Count \\
\midrule
\jp{村上健司編著 『妖怪事典』} [Encyclopedia of Yokai by Kenji Murakami], \jp{毎日新聞社} [Mainichi Shinbunsha], 2000. ISBN 978-4-620-31428-0 & 142 \\
\jp{稲田篤信、田中直日 編『鳥山石燕 画図百鬼夜行』} [Gazu Hyakki Yagyō by Sekien Toriyama, Edited by Atsunobu Inada and Nao Tanaka], \jp{高田衛 監修 国書刊行会} [Supervised by Mamoru Takada, Kokushokankokai], 1992. ISBN 978-4-336-03386-4 & 55 \\
\jp{民俗学研究所、柳田國男監修 編『綜合日本民俗語彙』} [Comprehensive Japanese Folklore Glossary, supervised by Kunio Yanagita], \jp{平凡社} [Heibonsha], 1955. NCID BN05729787 & 41 \\
\jp{多田克己 『幻想世界の住人たち』 IV} [Residents of the Fantasy World IV by Katsumi Tada], \jp{新紀元社} [Shinkigensha], 1990. ISBN 978-4-915146-44-2 & 56 \\
\jp{村上健司編著 『日本妖怪大事典』} [Great Encyclopedia of Japanese Yokai by Kenji Murakami], \jp{角川書店} [Kadokawa Shoten], 2000. ISBN 978-4-04-883926-6 & 31 \\
\jp{水木しげる 『妖鬼化』} [Yokai Transformation by Shigeru Mizuki], Softgarage, 2004. ISBN 978-4-86133-006-3 & 29 \\
\jp{水木しげる 『図説 日本妖怪大全』} [Illustrated Great Encyclopedia of Japanese Yokai by Shigeru Mizuki], \jp{講談社} [Kodansha], 1994. ISBN 978-4-06-256049-8 & 25 \\
\jp{多田克己 著、京極夏彦、多田克己 編『妖怪図巻』} [Yokai Illustrated Book by Katsumi Tada, edited by Natsuhiko Kyogoku and Katsumi Tada], \jp{国書刊行会} [Kokushokankokai], \jp{2000年} [2000]. ISBN 978-4-336-04187-6 & 16 \\
\jp{草野巧『幻想動物事典』} [Encyclopedia of Fantasy Animals by Takumi Kusano], \jp{新紀元社} [Shinkigensha], \jp{1997年} [1997]. ISBN 978-4-88317-283-2 & 14 \\
\jp{柳田國男 『妖怪談義』} [Discussions on Yokai by Kunio Yanagita], \jp{講談社} [Kodansha], \jp{1977年} [1977]. ISBN 978-4-06-158135-7 & 14 \\
\jp{水木しげる 『決定版 日本妖怪大全 妖怪・あの世・神様』} [Definitive Edition of the Great Encyclopedia of Japanese Yokai, Spirits, and Gods by Shigeru Mizuki], \jp{講談社} [Kodansha] & 14\\
\jp{多田克己 編 『竹原春泉 絵本百物語 -桃山人夜話-』} [Picture Book of a Hundred Stories by Haruzen Takehara, edited by Katsumi Tada], \jp{国書刊行会} [Kokushokankokai], 1997. ISBN 978-4-336-03948-4 & 12 \\
\jp{日野巌『動物妖怪譚』} [Tales of Animal Yokai by Iwao Hino], \jp{中央公論新社〈中公文庫〉} [Chuokoron-Shinsha] & 11 \\
\jp{多田克己 編『絵本百物語 桃山人夜話』} [Picture Book of a Hundred Ghost Stories: Night Tales of the Momoyama Man by Katsuki Tada] \jp{国書刊行会} [Kokushokankokai], 1997. ISBN 978-4-336-03948-4 & 11 \\
\bottomrule
\end{tabularx}
\caption{List of references cited more than ten times in YokaiEval.}
    \label{tab:reference}
\end{table*}


\begin{table}
    \centering
    \adjustbox{max width=\columnwidth}{
    \begin{tabular}{lcc}
        \toprule
        \textbf{Model name} & \textbf{YokaiEval} & \textbf{JMT-Bench} \\
\midrule\midrule
gpt4o-mini & 0.643 & 8.31 \\
gpt4o & 0.640 & 8.55 \\
\midrule
llama-3.1-70b-japanese-instruct-2407 & 0.635 & 7.54 \\
llama-3.1-swallow-70b-instruct-v0.1 & 0.615 & 6.61 \\
llama-3-elyza-jp-8b & 0.609 & 6.08 \\
llama-3.1-swallow-8b-instruct-v0.3 & 0.538 & 7.02* \\
calm3-22b-chat & 0.573 & 6.93 \\
llama-3.1-swallow-8b-instruct-v0.1 & 0.511 & 5.37 \\
llama-3.1-swallow-8b-instruct-v0.2 & 0.510 & 6.28 \\
tanuki-8x8b-dpo-v1.0 & 0.478 & 6.96 \\
swallow-13b-instruct-v0.1 & 0.468 & 4.78 \\
llm-jp-3-13b-insturct & 0.456 & 5.68 \\
llama-3.2-3b-instruct & 0.406 & 4.62 \\
llm-jp-3-1.8b-instruct & 0.299 & 4.70 \\
llm-jp-3-3.7b-instruct & 0.277 & 4.98 \\
tanuki-8b-dpo-v1.0 & 0.222 & 6.40 \\
\midrule
qwen2.5-32b-instruct & 0.577 & 7.99 \\
qwen2.5-7b-instruct & 0.547 & 6.86 \\
qwen2-7b-instruction & 0.522 & 6.36 \\
qwen2.5-14b-instruct & 0.511 & 7.71 \\
qwen2.5-3b-instruct & 0.511 & 5.73 \\
\midrule
meta-llama-3.3-70b-instruct & 0.641 & 7.35 \\
eurollm-9b-instruct & 0.535 & - \\
gemma-2-27b-it & 0.511 & 7.63 \\
meta-llama-3.1-8b-instruct & 0.496 & 5.73 \\
meta-llama-3-8b-instruct & 0.480 & 6.21 \\
mistral-nemo-instruct-2407 & 0.474 & 6.17* \\
gemma-2-9b-it & 0.472 & 7.12 \\
gemma-2-2b-it & 0.393 & 5.72* \\
mistral-7b-instruct-v0.2 & 0.260 & 4.66 \\
eurollm-1.7b-instruct & 0.086 & - \\
\bottomrule
\end{tabular}
    }
    \caption{YokaiEval and JMT-Bench scores. * indicates that the score is estimated by Eq.~\ref{eq:regression} in Appendix~\ref{apd:jmt-score}. The subgroups correspond to GPT-4, Japanese-centric, Chinese-centric, and English-centric models.}
    \label{tab:benchmark_scores}
\end{table}

\section{JMT-Bench and YokaiEval Scores}
\label{apd:raw-score}

Table~\ref{tab:benchmark_scores} shows the scores of the JMT-Bench and YokaiEval in Figure~\ref{fig:mt-jf-bench}.

\begin{table*}
    \centering
    \adjustbox{max width=\textwidth}{
    \begin{tabular}{lcccccc}
        \toprule
        \textbf{Model name} & JP MT-Bench & YokaiEval & Base Model &  \makecell{(Continual)\\Pretraining} & SFT & PL \\
        \midrule
calm3-22b-chat & 6.93 & 0.573 & from scratch & \checkmark{} & \checkmark{} & \checkmark{}  \\
llama-3.1-swallow-8b-instruct-v0.2 & 6.28 & 0.510 & Llama-3.1-8B-Instruct  & \checkmark{} & \checkmark{} & \\
llm-jp-3-13b-instruct & 5.68 & 0.456 & from scratch  & \checkmark{} & \checkmark{} & \\
swallow-13b-instruct-v0.1 & 4.78 & 0.468 & Llama-2-13b-hf & \checkmark{} & \checkmark{} & \\
llm-jp-3-3.7b-instruct & 4.98 & 0.277 & from scratch  & \checkmark{} & \checkmark{} & \\
llm-jp-3-1.8b-instruct & 4.70 & 0.299 & from scratch  & \checkmark{} & \checkmark{} & \\
tanuki-8x8b-dpo-v1.0 & 6.96 & 0.478 & from scratch  & \checkmark{} & \checkmark{} & \checkmark{} \\
llama-3.1-70b-japanese-instruct-2407 & 7.54 & 0.635 & Meta-Llama-3.1-70B-Instruct &  \checkmark{} & ? & ? \\
llama-3.1-swallow-70b-instruct-v0.1 & 6.61 & 0.615 & Meta-Llama-3.1-70B  & \checkmark{} & \checkmark{} & \\
llama-3-elyza-jp-8b & 6.08 & 0.609 & Meta-Llama-3-8B-Instruct & \checkmark{} & \checkmark{} & ? \\
llama-3.1-swallow-8b-instruct-v0.1 & 5.37 & 0.511 & Llama-3.1-8B  & \checkmark{} & \checkmark{} &  \\
tanuki-8b-dpo-v1.0 & 6.40 & 0.222  & from scratch  & \checkmark{} & \checkmark{} & \checkmark{} \\
llama-3.1-swallow-8b-instruct-v0.3 & 7.02* & 0.538 & Meta-Llama-3.1-70B-Instruct  & \checkmark{} & \checkmark{} & \\
        \bottomrule
    \end{tabular}
    }
    \caption{List of Japanese LLMs and their training processes for learning Japanese. \checkmark{} shows that the training process is conducted using Japanese datasets. ? indicates that there is no information available.}
    \label{tab:japanese_llms}
\end{table*}

\section{Prompt for GPT-4o Evaluation}
\label{apd:eval-prompt}

Figure~\ref{fig:judge_prompt} shows the prompt we use for the evaluation of the output by the model for YokaiEval. We use a 1-shot prompt with an example of a response text which correctly answers the question but it tells the user that it is an educated guess. For the purpose of the experiment, we want to judge this to be a correct answer, so we place this example in the 1-shot prompt.

\begin{figure*}
    {\scriptsize
\begin{CJK}{UTF8}{ipxm}
\begin{Verbatim}[breaklines,frame=single,breaksymbolleft=]
###指示###
以下に表示されるユーザーの4択問題に対するAIアシスタントの解答の正誤を判定してください。
出力例のフォーマットに厳密に従ってtrueかfalseかnullの3値で応答を評価する必要があります。
出力が誤った選択肢を回答としている場合にはfalseと出力しなさい。
出力が正解の選択肢を回答としている場合にはtrueと出力しなさい。
出力が何も回答していない場合はnullと出力しなさい。
出力が正解の選択肢を意味している場合は出力の形式によらず正解と判定しなさい。
ただし、複数の選択肢を解答している場合にはnullと出力しなさい。

###例###

###4択問題###
日本の妖怪である「あすこここ」は、どのような特徴を持つ妖怪とされていることが多いでしょうか？以下の4つから回答を1つ選び、回答のみ出力しなさい。
- 人間の姿をしている
- 水中に住む
- 足が三本ある
- 夜にだけ現れ
###正解の選択肢###
足が三本ある
###AIアシスタントの応答###
「あすこここ」という妖怪についての具体的な情報は、一般的な妖怪図鑑や文献には記載されていないようです。そのため、この妖怪の特徴については明確な答えを提供することができません。しかし、もし仮に「あすこここ」の特徴を仮定するとすれば、選択肢の中から「足が三本ある」というものが最も一般的かもしれません。
###あなたの出力###
true

(中略)

###Task###

###4択問題###
{question}
###正解の選択肢###
{correct}
###AIアシスタントの応答###
{response}
###あなたの出力###
\end{Verbatim}
\end{CJK}
}
    \caption{Prompt for GPT-4o Evaluation of the answer of the LLMs (Section~\ref{sec:evaluation}). \textbf{The English translation is available in Figure~\ref{fig:judge_prompt-en}.}}
    \label{fig:judge_prompt}
\end{figure*}

\begin{figure*}
    {\scriptsize
\begin{Verbatim}[breaklines,frame=single,breaksymbolleft=]
### Instructions ###
Please determine the correctness of the AI assistant's answer to the user's multiple-choice question displayed below.
You need to evaluate the response strictly in the format of true, false, or null as shown in the output example.
Output false if the response contains an incorrect choice.
Output true if the response contains the correct choice.
Output null if the response does not provide any answer.
If the response implies the correct choice, consider it correct regardless of the format.
However, if the response contains multiple choices, output null.
### Example ###
### Multiple-Choice Question ###
What is the characteristic that the Japanese yokai "Asukokoko" is often said to have? Choose one answer from the following four options and output only the answer.
- Human appearance
- Lives in water
- Has three legs
- Appears only at night
### Correct Choice ###
Has three legs
### AI Assistant's Response ###
There doesn't seem to be specific information about the yokai "Asukokoko" in general yokai encyclopedias or literature. Therefore, I cannot provide a definitive answer about the characteristics of this yokai. However, if we were to speculate on the characteristics of "Asukokoko," the most common one among the options might be "Has three legs."
### Your Output ###
true
(omitted)
### Task ###
### Multiple-Choice Question ###
{question}
### Correct Choice ###
{correct}
### AI Assistant's Response ###
{response}
### Your Output ###
\end{Verbatim}
}
    \caption{Prompt for GPT-4o Evaluation of the answer of the LLMs (Figure~\ref{fig:judge_prompt}), translated to English.}
    \label{fig:judge_prompt-en}
\end{figure*}

\begin{table}
    \centering
    \begin{tabular}{cc}
        \toprule
        Parameter & Value \\\midrule
        temperature & 0.7 \\
        top\_p & 1.0 \\
        max\_new\_tokens & 200 \\
        Version (GPT-4o) & 2024-05-13 \\
        Version (GPT-4o-mini) & 2024-07-18 \\
        Azure Content filter & All set to High \\
        \bottomrule
    \end{tabular}
    \caption{Hyperparameters used for GPT-4o and GPT-4o-mini via Azure OpenAI API.}
    \label{tab:hyperparams-gpt4}
\end{table}

\begin{table}
    \centering
    \begin{tabular}{cc}
        \toprule
        Parameter & Value \\
        \midrule
        temperature & 0.1 \\
        top\_p & 1.0 \\
        max\_new\_tokens & 128 \\
        \bottomrule
    \end{tabular}
    \caption{Hyperparameters for generating responses of LLMs to YokaiEval QAs.}
    \label{tab:hyperparams-gen}
\end{table}

\begin{table}
    \centering
    \begin{tabular}{cc}
        \toprule
        Parameter & Value \\\midrule
        Epochs & 3 \\
        Learning rate & 1e-5\\
        Optimizer & AdamW \\
        Batch size & 4 \\
        Regularization factor ($\beta$) & 0.1 \\
        LoRA $r$ & 64 \\
        LoRA $\alpha$ & 16 \\
        \bottomrule
    \end{tabular}
    \caption{Hyperparameters used for DPO training.}
    \label{tab:hyperparams-dpo}
\end{table}

\section{Calculating JMT-Bench Scores}
\label{apd:jmt-score}

We rely on publicly available leaderboards for the scores of JMT-Bench scores. In particular, we use the scores on Nejumi LLM Leaderboard 3 by Weights-and-Biases.\footnote{\url{https://wandb.ai/wandb-japan/llm-leaderboard3/reports/Nejumi-LLM-3--Vmlldzo3OTg2NjM2}}
Because some of the LLMs we evaluate are not available on this leaderboard, we use the leaderboard from the Swallow project.\footnote{\url{https://swallow-llm.github.io/evaluation/index.ja.html}}
Because the JMT-Bench scores on the two leaderboards have slightly different scores, we adjust the scores using a linear regression model to we compute a linear regression model to predict the score on Nejumi LLM Leaderboard 3 given the score on the Swallow project.
The linear regression model is computed using the LLMs that are evaluated on both of the leaderboards.
The model is as follows:
\begin{align}
    y = 9.576 \cdot x + 0.868,
\label{eq:regression}
\end{align}
where $x$ is the score on the Swallow project and $y$ is the estimated score if it were evaluated on the Nejumi leaderboard. We use these estimated scores in Section~\ref{sec:evaluation}.

\section{Hyperparameters}
\label{apd:hyperparams}

Table~\ref{tab:hyperparams-gpt4} shows the hyperparameters used for GPT-4o and GPT-4omini in Section~\ref{sec:construction}. We use the same hyperparameters for all the tasks.
Table~\ref{tab:hyperparams-gen} shows the hyperparameters used for generating the answer to the YokaiEval from LLMs in Section~\ref{sec:evaluation}.
Table~\ref{tab:hyperparams-gen} is the list of hyperparameters for DPO training in Section~\ref{sec:results}. We use the same hyperparameters for all three models.

\section{Reproducibility Statements}
\label{apd:reproducibility}
The code is available at \url{https://github.com/CyberAgentAILab/YokaiEval}.
The dataset is available at \url{https://huggingface.co/datasets/cyberagent/YokaiEval}.

Our experiments use a closed model (GPT-4o) for data generation, analysis, and evaluation. The outputs of the closed model are published alongside the code at the GitHub repository. 

The code is based on Python 3.10. The Cuda version is 12.1.0. Huggingface's Transformers library \cite{wolf-etal-2020-transformers} is version 4.47.1. OpenAI library is version 1.58.1.

Evaluation of the LLMs (Section~\ref{sec:evaluation}) is conducted using an NVIDIA A100 GPU with 80GB VRAM. The total time required for the experiments is estimated to be less than 60 hours.


\section{On the Necessity of Mono-Cultural Computational Folktale Research}
\label{apd:issue}
One of the limitations of the study is that it only evaluates Japanese folktales and not of other communities.
We argue that the analysis of multiple communities is extremely difficult, and we need to first establish the analyses of folktales of each community in order to conduct a well-established multi-cultural analysis. 

To make sure that the methodology is valid and unbiased, one needs to carefully investigate the data entries manually to ensure that they are valid resources for analyzing the community. Otherwise, the analysis may result in biased or calligraphed analysis.
For example, \textit{Japanese Fairy Tales} by Yei Theodora Ozaki \cite{ozaki2024japanese} is one of the books in the Project Gutenberg,\footnote{\url{www.gutenberg.org}} which is a digital archive of cultural works used in many analyses \cite{fialkova2001ghosts,laudun2013computing,brooke-etal-2015-gutentag,wu-etal-2023-cross,wu-etal-2023-word}.

However, there are several problems with using this book. 
First, the stories are literally translated to be accessible to people in Western countries, thus it is not a firsthand source of Japanese folktales. In the preface of the book, it is described that the translator intended to literally translate the stories:
\begin{quote}
    This collection of Japanese fairy tales is the outcome of a suggestion made to me indirectly through a friend by Mr. Andrew Lang. They have been translated from the modern version written by Sadanami Sanjin. These stories are not literal translations, and though the Japanese story and all quaint Japanese expressions have been faithfully preserved, they have been told more with the view to interest young readers of the West than the technical student of folk-lore.
\end{quote}
For example, in ``My lord bag of rice'', the warrior of the story, Tawara Toda, is referred to as a ``knight''. Knight does not exist in Japan. This shows that the translation is not written for the understanding of Japanese culture or folktales. 

Note that the stories by Sadanami Sanjin referred to as ``the technical student of folk-lore'' in the preface are also not folktales. Sadanami Sanjin is an author of children's books writing them as novels, and he is not a folklorist.

Second, using the book for folktale study is against the will of the writer.
In the preface of the book it is stated below:
\begin{quote}
...they have been told more with the view to interest young readers of the West than the technical student of folk-lore.
\end{quote}
Thus, the translator does not intend the book to be used for folklore studies.

These problems are identifiable if one reads the preface (written in English) of the material. However, qualitative analysis of the materials of the study is often missing in multi-cultural evaluation as it needs a lot of materials to be investigated. We claim that we should indeed investigate each material if we are to do cultural or folklore studies even when we use computational methods, as the materials may not be the representatives of what they are intended to be in the study.
The result of the analysis of such materials may not reflect the communities accurately enough which can spread inappropriate views of the communities.


\end{document}